\documentclass{article} 
\usepackage{collas2025_conference,times}
\usepackage{easyReview}

\usepackage{amsmath,amsfonts,bm}

\def\eqref#1{equation~\ref{#1}}

\def\1{\bm{1}}

\DeclareMathAlphabet{\mathsfit}{\encodingdefault}{\sfdefault}{m}{sl}
\SetMathAlphabet{\mathsfit}{bold}{\encodingdefault}{\sfdefault}{bx}{n}

\DeclareMathOperator*{\argmin}{arg\,min}

\usepackage{hyperref}
\hypersetup{
    colorlinks=true,
    linkcolor=red,
    filecolor=magenta,
    urlcolor=blue,
    citecolor=purple,
    pdftitle={Overleaf Example},
    pdfpagemode=FullScreen,
    }
\usepackage{graphicx}
\usepackage{amsmath}
\usepackage{amssymb}
\usepackage{subcaption}
\usepackage{url}
\usepackage{algorithm}
\usepackage{algorithmic}
\usepackage{multirow}

\DeclareMathOperator{\arctanh}{arctanh}
\newcommand{\Lagr}{\mathcal{L}}

\definecolor{Gray}{HTML}{C8E8FA}
\usepackage{wrapfig}
\usepackage{xcolor, colortbl}
\usepackage{tabularx}
\usepackage{booktabs}

\newcommand{\reb}[1]{\textcolor{black}{#1}}
\newcommand{\mymethod}{HyperCLIC }
\newcommand{\rebuttal}[1]{\textcolor{black}{#1}}
\usepackage[normalem]{ulem}

\title{Continual Hyperbolic Learning of \\Instances \texttt{and} Classes}

\author{Melika Ayoughi \\
University of Amsterdam, \\
\texttt{m.ayoughi@uva.nl} \\
\AND 
Mina GhadimiAtigh\thanks{Equal contribution.}  \\
University of Amsterdam, \\
\texttt{m.ghadimiatigh@uva.nl} \\
\And 
Mohammad Mahdi Derakhshani\footnotemark[1] \\
University of Amsterdam, \\
\texttt{m.m.derakhshani@uva.nl} \\
\And
Cees Snoek \\
University of Amsterdam, \\
\texttt{c.g.m.snoek@uva.nl} \\
\And
Pascal Mettes \\
University of Amsterdam, \\
\texttt{p.s.m.mettes@uva.nl} \\
\And
Paul Groth \\
University of Amsterdam, \\
\texttt{p.t.groth@uva.nl}}

\preprintcopy 

\begin{document}

\maketitle
\begin{abstract}
Continual learning has traditionally focused on classifying either instances or classes, but real-world applications, such as robotics and self-driving cars, require models to handle both simultaneously. To mirror real-life scenarios, we introduce the task of continual learning of instances \textit{and} classes, at the same time. This task challenges models to adapt to multiple levels of granularity over time, which requires balancing fine-grained instance recognition with coarse-grained class generalization. In this paper, we identify that classes and instances naturally form a hierarchical structure. To model these hierarchical relationships, we propose HyperCLIC, a continual learning algorithm that leverages hyperbolic space, which is uniquely suited for hierarchical data due to its ability to represent tree-like structures with low distortion and compact embeddings. Our framework incorporates hyperbolic classification and distillation objectives, enabling the continual embedding of hierarchical relations. To evaluate performance across multiple granularities, we introduce continual hierarchical metrics. We validate our approach on EgoObjects, the only dataset that captures the complexity of hierarchical object recognition in dynamic real-world environments. Empirical results show that HyperCLIC operates effectively at multiple granularities with improved hierarchical generalization.

\end{abstract}
\section{Introduction}

Continual learning addresses a long-standing challenge in machine learning: learning from new classes often leads to catastrophic forgetting of old classes \citep{ewc, bic, magistri2024empirical, lyle2024disentangling, delange2021continualsurvey, wang2023comprehensive}. To mitigate this, numerous solutions have been proposed, including data replay \citep{bang2021rainbow, wang2021ordisco}, regularization \citep{yin2021mitigating, lee2020continual}, and knowledge distillation \citep{afc, glfc}. While these methods primarily focus on class-level discrimination, fewer works have expanded the scope to instance-level continual learning. In robotics, for example, classifying specific instances of objects enables robots to make informed decisions about their use or placement \citep{ammirato2019recognizing, singh2014bigbird, held2016robust}. Several benchmarks have been developed to recognize instances under varying conditions such as illumination, occlusion, or background \citep{borji2016ilab, core50, openloris}. \cite{continual_instances} and \cite{kilickaya2023labels} address incremental instance recognition using metric learning under an object re-identification setting and self-supervised learning respectively. The recent EgoObjects \citep{egoobjects} benchmark and its instance-level continual challenge \citep{clvision_challenge} further highlight the complexity, scale and open-ended nature of instance-level recognition in dynamic real-world environments.

In real-life scenarios, recognition at a \textit{single} granularity level is often insufficient. For critical domains such as robotics, self-driving cars, and medical imaging, it is crucial that even if instance-level predictions are incorrect, class-level predictions remain accurate to prevent catastrophic outcomes and enable quick generalization to new instances. This is challenging because class-level continual learning, by design, is invariant to specific instances, while instance-level continual learning without class-level awareness risks significant errors.
We introduce the problem of \textit{joint} instance- and class-level continual learning, which aims to perform recognition simultaneously at both levels of granularity.

\begin{wrapfigure}{r}{0.52\textwidth} 
\centering
\vspace{-0.2cm}
\includegraphics[width=\linewidth, trim={45cm 8cm 45cm 8cm}, clip]{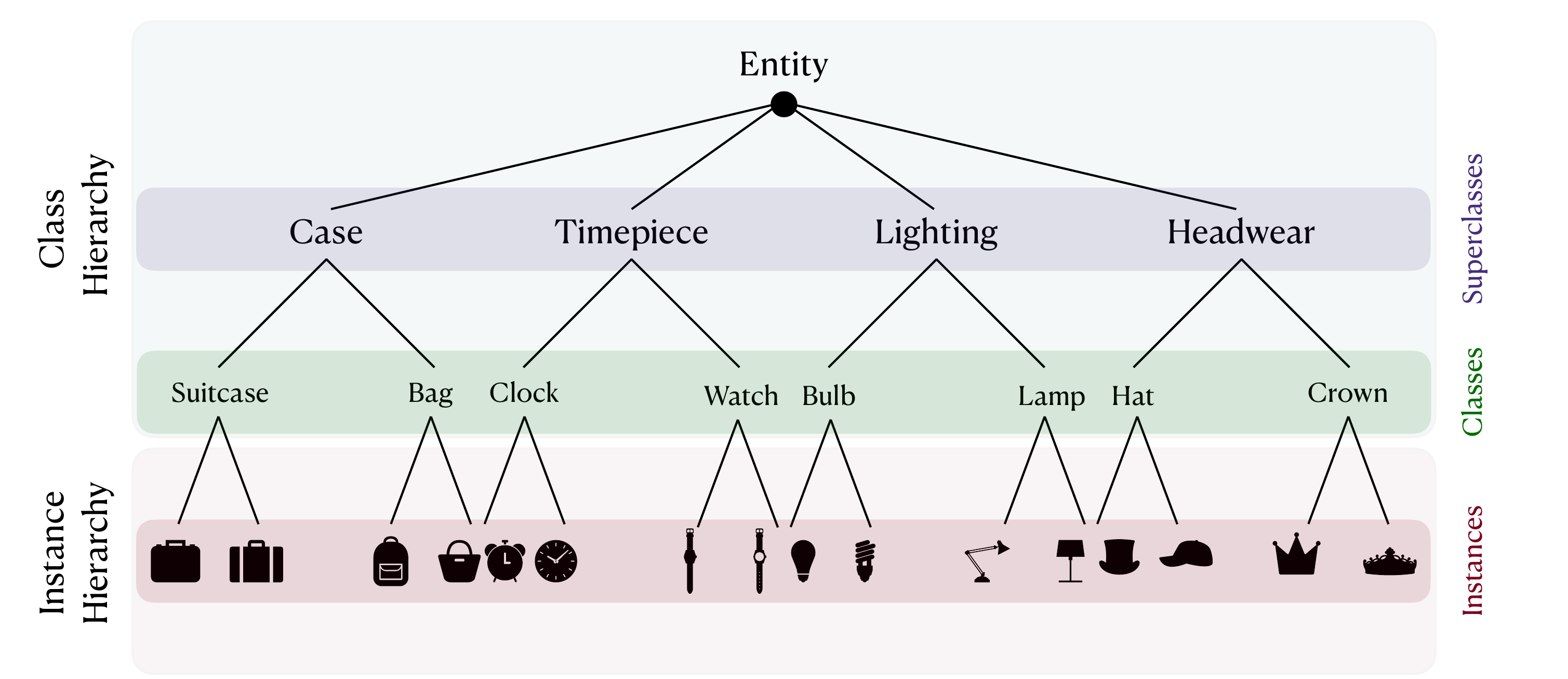}
\caption{\textbf{Recognizing instances and classes simultaneously} is important in many real-world applications. By adding instances as an additional layer to the object hierarchy and learning representations that capture the joint hierarchy, we can classify samples at multiple levels of granularity.}
\label{fig:hierarchy}
\vspace{-0.2cm}
\end{wrapfigure}

To tackle this challenge, we leverage the inherent hierarchical structure of instances and classes. Classes form a coarse-to-fine hierarchy \citep{wordnet, imagenet}, and instances add an additional layer to this hierarchy \citep{yan2024concept}. Figure \ref{fig:hierarchy} illustrates this hierarchical organization. \rebuttal{Theoretically, hyperbolic space is known for its ability to embed tree-structured data with minimal distortion \citep{sarkar2011low}. This property stems from two fundamental characteristics of hyperbolic geometry: negative curvature and exponential volume growth. In contrast to Euclidean space, where polynomial volume growth leads to inevitable distortion when embedding branching in high dimensions, hyperbolic space naturally accommodates hierarchical representations. This geometric advantage precisely matches the exponential properties in real-world taxonomies. Thus, hyperbolic learning has been used in many hierarchical computer vision tasks \citep{ibrahimi2024intriguing, desai2023hyperbolic, mettes2024hyperbolic, ge2023hyperbolic, hong2023hyperbolic}. Due to the strong hierarchical nature of the problem of joint class- and instance-level continual learning, hyperbolic learning provides a natural solution for preserving such relationships during incremental learning.} We propose HyperCLIC, a hyperbolic continual learner that embeds this hierarchy in hyperbolic space for \textit{joint} instance- and class-level recognition. Our contributions are threefold: (i) we formally define the task of continual learning of instances and classes and introduce continual hierarchical metrics to measure accuracy at both levels and evaluate the severity of hierarchical mistakes; (ii) we introduce HyperCLIC, which employs hyperbolic classification and distillation losses to enable incremental learning while preserving hierarchical relationships; and (iii) we demonstrate the effectiveness of our approach on the EgoObjects benchmark, achieving strong performance across multiple granularities and making hierarchically better mistakes.


\section{Literature Review}

\paragraph{Continual Learning}
Continual learning is primarily studied under three main scenarios \citep{van2022three, chen2018lifelong}: task-incremental, class-incremental, and domain-incremental learning. In task-incremental learning \citep{van2022three, ewc, til_li2017learning}, a model is trained on incremental tasks with clear boundaries. The task ID is known during test time. In contrast, in the class-incremental learning \citep{cil_guo2022online, cil_kim2022theoretical, cil_kim2023learnability, bic, icarl}, the task ID is not provided \rebuttal{\sout{, making task-incremental a particular case of class-incremental}}. Domain-incremental \citep{shi2024unified, dil_garg2022multi,dil_kalb2021continual, dil_mirza2022efficient, dil_wang2022s} focuses on scenarios where the data distribution's incremental shift is explicitly modeled. Here, we focus on class-incremental learning. 
Regardless of continual scenarios, the main challenge of continual learning is to alleviate catastrophic forgetting with only limited access to the previous data \citep{evron2022catastrophic, shi2021overcoming}. Catastrophic forgetting means that performance on previously learned tasks degrades significantly when learning new tasks. To address this, various strategies have been developed. \cite{CI_survey} groups class-incremental methods into data-centric, model-centric, and algorithm-centric. Data-centric methods concentrate on solving class-incremental learning with exemplars by using data replay \citep{prabhu2020gdumb, bang2021rainbow, chaudhry2018riemannian, aljundi2019gradient, rolnick2019experience, shin2017continual, wang2021ordisco} or data regularization techniques \citep{lopez2017gradient, chaudhry2018efficient, zeng2019continual}. Model-centric methods either regularize the model parameters from drifting away \citep{ewc, zenke2017continual, yin2021mitigating, lee2020continual} or dynamically expand the network structure for stronger representation ability \citep{der, memo, wang2022foster, douillard2022dytox}. Algorithm-centric methods either utilize \rebuttal{knowledge distillation \citep{hinton2015distilling}} to resist forgetting \citep{icarl, afc, til_li2017learning, lucir, der++, glfc, podnet} or rectify the bias in the model \citep{bic, il2m, fact, wa}. \rebuttal{Several recent works, such as \cite{lee2023online, wang2022hierarchical, muralidhara2024cleo, tran2024leveraging} highlight the importance of hierarchies in continual learning by exploring hierarchical task ordering configurations, our approach differs fundamentally: we treat hierarchy as prior knowledge for geometric representation learning rather than as a task configuration mechanism. These methods focus on ordering the tasks to first learn coarse-grained classes, with data labels continually expanding to more fine-grained classes. This distinction positions our work as complementary to - rather than competitive with - methods that modify the task sequence.} 


\paragraph{Hyperbolic Learning}
Hyperbolic learning has gained considerable attention in deep learning in embedding taxonomies and tree-like structures~\citep{poincare, entailment, law2019lorentzian, nickel2018learning}, graphs~\citep{liu2019hyperbolic, chami2019hyperbolic, bachmann2020constant, dai2021hyperbolic}, and text~\citep{tifrea2018poincar, zhu2020hypertext, dhingra2018embedding, leimeister2018skip}. Hyperbolic space is a space with constant negative curvature that can be thought of as a continuous version of a tree, making it a good choice to embed any finite tree while preserving the distances~\citep{ungar2008gyrovector, hamann2018tree}. Based on the tree-like behavior,~\cite{poincare} introduce a new approach to embed symbolic data in the Poincar\'e ball model, a particular hyperbolic space model.~\cite{entailment} take a step forward and improve the Poincar\'e embeddings~\citep{poincare} using a model based on the geodesically convex entailment cones, showing the effectiveness when embedding data with a hierarchical structure. Furthermore, hyperbolic space has also been used to develop intermediate layers~\citep{entailment, cho2019large}, and deep neural networks~\citep{ganea2018hyperbolic, shimizu2020hyperbolic}.

Following the initial success, hyperbolic space has shown success and gained attention in computer vision tasks in supervised and unsupervised learning~\citep{mettes2024hyperbolic}. Hyperbolic embeddings have shown benefits in classification and few-shot learning~\citep{khrulkov2020hyperbolic, gao2021curvature, guo2022clipped, ghadimi2021hyperbolic}, zero-shot learning~\citep{liu2020hyperbolic, hong2023hyperbolic}, segmentation~\citep{atigh2022hyperbolic, chen2023hyperbolic}, out-of-distribution generalization~\citep{ganea2018hyperbolic, van2023poincare}, uncertainty quantification~\citep{atigh2022hyperbolic, chen2023hyperbolic}, contrastive learning~\citep{yue2023hyperbolic, ge2023hyperbolic}, hierarchical representation learning~\citep{actionsonhyperbole, dhall2020hierarchical, doorenbos2024hyperbolic}, generative learning~\citep{cho2024hyperbolic, dai2020apo, mathieu2020riemannian}, and vision-language representation learning~\citep{ibrahimiintriguing, desai2023hyperbolic}. To our knowledge, \cite{gao2023exploring} is the only study that utilizes non-Euclidean geometries for class-level continual learning. Their approach introduces an expanding geometry of mixed-curvature space, targeting low-memory scenarios. We propose hyperbolic continual learning tailored for joint learning of instances and classes in a default memory regime.

\section{Continual learning of instances and classes}

\paragraph{Task definition}
We follow a class-incremental learning setup, where new instance classes are introduced at each new task. Consider a sequence of $T$ training tasks $\mathcal{D}^1, \mathcal{D}^2, ..., \mathcal{D}^T$ with non-overlapping instances, where $\mathcal{D}^t = {(x_i^t, y_i^t)}_{i=1}^{n_t}$ is the $t$-th incremental step with $n_t$ training samples. Each $x_i^t \in \mathbb{R}^{\mathcal{D}}$  is an example of instance $y_i \in Y_t$, where $Y_t$ is the label space of task $t$. We formalize the hierarchy of instances, classes, superclasses, and other nodes as a tree $\mathcal{T}=(V,E)$. Each label in $Y_t$ corresponds to a leaf node label of tree $\mathcal{T}$, as detailed in tree definition. We only have access to samples of instances in $\mathcal{D}^t$ when training task $t$.

\paragraph{Tree definition} The joint class-instance hierarchy $\mathcal{T}=(V,E)$ (Figure \ref{fig:hierarchy}) consists of four types of vertices $V=V_I \cup V_C \cup V_S \cup V_O$ and four types of directed edges $E= E_{IC} \cup E_{CS} \cup E_{SO} \cup E_{OO}$. Vertices $V_I=\{i_1,i_2,...,i_{n_I}\}$ are the set of instances, $V_C=\{c_1, c_2, ..., c_{n_C}\}$ the set of classes that are parents of instances, $V_S=\{s_1,s_2,..., s_{n_S}\}$ the set of superclasses that are parents of classes, and $V_O=\{o_1,o_2,..., o_{n_O}\}$ are the set of other remaining nodes. Each node $v \in V$ corresponds to a distinct label. The number of nodes $|V|$ equals the number of all labels $|C|$ in the hierarchy. $E_{IC} = \{(i_j, c_k) \mid i_j \in V_I, c_k \in V_C\} $ represent edges between instances and classes, $E_{CS} = \{(c_k, s_m) \mid c_k \in V_C, s_m \in V_S\}$ edges between classes and superclasses, $E_{SO} = \{(s_m, o_p) \mid s_m \in V_S, o_p \in V_O\}$ edges between superclasses and other nodes,
and $E_{OO} = \{(o_p, o_q) \mid o_p \in V_O, o_q \in V_O\}$ represents the remaining 1-hop hypernymy relations. \rebuttal{In other words, $E_{oo}$ represents all other remaining edges of the tree between nodes $A$ and $B$, where neither A or B are instance ($V_I$), class ($V_C$), or superclass ($V_S$) nodes.} This formalization defines the hierarchical relationships in a tree structure where each node $V$ (instance, class, superclass, and other) except the root has one parent. Each instance $V_I$ is a leaf node in the tree structure $\mathcal{T}$.

\paragraph{Continual hierarchical evaluation}
Evaluation is done using both standard and continual hierarchical metrics. The standard metrics include average forgetting, average accuracy, and per-task accuracies. In our work, the accuracy metric reflects the instance-level accuracy, so we refer to it as the continual instance-level accuracy. The instance-level accuracy on task $t$ after incremental learning of task $t'$ is defined as $\texttt{Acc}_{instance}^{tt'} = \frac{1}{n_t} \sum_{i=1}^{n_t} 1\{\hat{y_i}^{tt'}=y_i^t\}$, where $\hat{y_i}^{tt'}$ is the predicted class for instance label $y_i^t$ evaluated on the test set of $t$-th task after incremental learning of the $t'$-th task. To evaluate continual hierarchical consistency and robustness, we report continual class-level and superclass-level accuracies, inspired by sibling and cousin accuracy in \cite{ghadimi2021hyperbolic}. For each sample $x_i^t$ and its ground-truth instance label $y_i^t \in V_I$, let $p(y_i^t)$ be the parent class $y_i^t$ and $gp(y_i^t)$ be the grandparent class of $y_i^t$. In class-level accuracy, a prediction is also correct if it shares a parent with the target class. The class-level accuracy on task $t$ after incremental learning of task $t'$ is defined as $\texttt{Acc}_{class}^{tt'} = \frac{1}{n_t} \sum_{i=1}^{n_t} 1\{p(\hat{y_i}^{tt'})=p(y_i^t)\}$.
In the superclass-level accuracy, the predicted labels must share a grandparent with the target class to count as correct. The superclass-level accuracy on task $t$ after incremental learning of task $t'$ is defined as $\texttt{Acc}_{superclass}^{tt'} = \frac{1}{n_t} \sum_{i=1}^{n_t} 1\{gp(\hat{y_i}^{tt'})=gp(y_i^t)\}$. Similar to \cite{bertinetto2020making, garg2022learning}, we also report the continual distance to the Lowest Common Ancestor (LCA) that captures the mistake's hierarchical severity. For the wrong predictions, LCA distance reports the average number of edges between the predicted node and the LCA of the predicted and ground-truth nodes. LCA distance reveals how hierarchically far off the inaccurate predictions are, considering the joint class-instance hierarchy as the ground truth. Following the 3rd CLVision challenge \citep{verwimp2023clad} and the SSLAD competition \citep{clvision_challenge}), all continual hierarchical metrics are calculated as the average mean over all the tasks. Here, we take the continual instance-level accuracy as an example. Let $Acc_{instance}^{ij} \in [0, 1]$ denote the instance-level classification accuracy evaluated on the test set of the $i$-th task after incremental learning of the $j$-th task. The average mean instance-level accuracy is defined as $\texttt{Accuracy}_\texttt{instance-level} = \frac{1}{T^2}\sum_{j=1}^{T} \sum_{i=1}^{T} Acc_{instance}^{ij}$, where $T$ denotes the number of all tasks. We use the same formulation for average mean class-level accuracy and superclass-level accuracies. We also report the per-task accuracies at the end of the last task.

\section{\mymethod}
\begin{figure}
\centering
\includegraphics[width=\linewidth, trim={0cm 5cm 32cm 0cm}, clip]{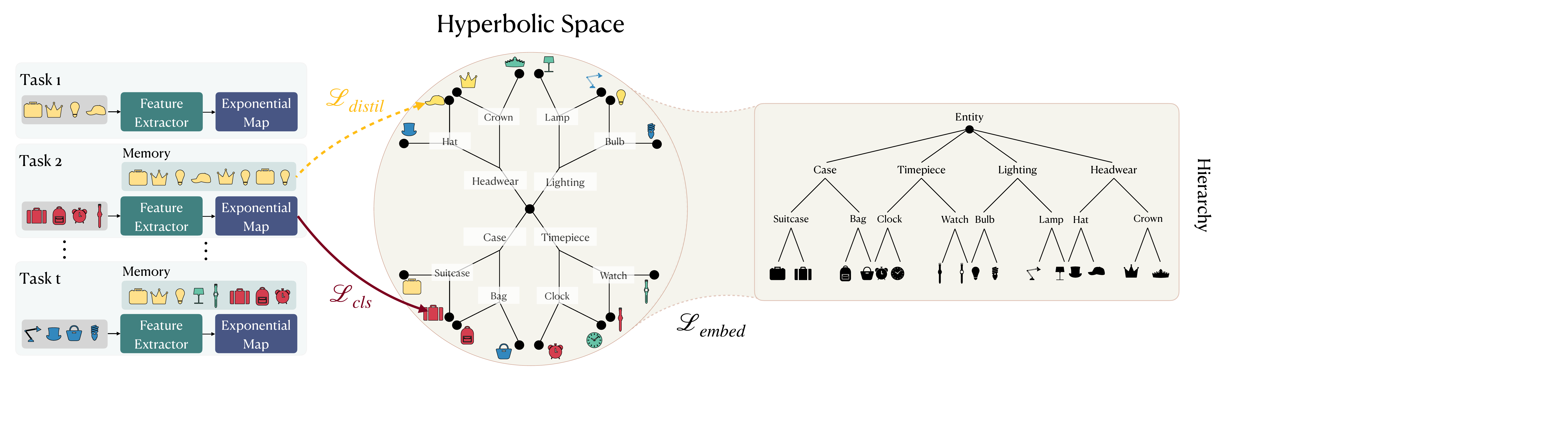}
\caption{\textbf{Overview of \mymethod.} The class-instance hierarchy is projected into a shared hyperbolic space. When learning an instance, its samples go through the feature extractor and are mapped into the shared hyperbolic space using the exponential map. These samples are then pushed toward their hyperbolic instance prototype via classification loss and are encouraged to maintain hyperbolic logits from previous classes through hyperbolic distillation.}
\label{fig:method}
\end{figure}

In this section, we present our method for continual learning of instances and classes using hyperbolic geometry. Our goal is to continually learn hierarchy-aware representations, enabling classification at different levels of granularity: instance-, class-, and superclass-level. Our method consists of two stages. The objective of the first stage (Section 4.1) is to obtain a set of hyperbolic prototypes (Algorithm \ref{alg:first_stage}), which are later used in the second stage (Section 4.2) for the classification loss (Algorithm \ref{alg:second_stage}). Figure \ref{fig:method} illustrates the two main components of HyperCLIC: First, we embed the class-instance hierarchy into hyperbolic space using Poincar\'e embeddings and entailment cones, leveraging hyperbolic space's ability to model hierarchical relationships. Second, we perform continual hyperbolic alignment between visual inputs and the embedded hierarchy. We apply a hyperbolic prototype-based loss for classifying new instances and a hyperbolic distillation loss to maintain the consistency of embeddings for previously seen instances, ensuring that the model respects the hierarchical structure during continual learning. Code is available at \href{https://anonymous.4open.science/r/HyperCLIC/}{this} link.

\subsection{Embedding class-instance hierarchies in hyperbolic space}
\label{sec:formalize_hierarchy}


Given the class-instance tree hierarchy $\mathcal{T}$, our goal is to embed the symbolic tree representation \emph{a priori} into a hyperbolic embedding space that incorporates the hierarchical relations between classes and from classes to instances.
We define the hyperbolic manifold using the Poincar\'e ball model \citep{poincare}, as it is well-suited for gradient-based optimization. More formally, let $\mathbb{B}_c^d = \{\mathrm{z} \in \mathbb{R}^d | c\|\mathrm{z}\|^2 < 1\}$ be the open d-dimensional unit ball, where $||.||$ denotes the Euclidean norm, and $c$ denotes the curvature. The Poincar\'e ball model corresponds to the Riemannian manifold $(\mathbb{B}_c^d, g_\mathrm{z}^{\mathbb{B}})$ with the Riemannian metric tensor $g_\mathrm{z}^{\mathbb{B}} = 4(1- c\| \mathrm{z}\|^2)^{-2}\mathbf{I}_d$.

Let $d_{\mathcal{T}}(v_i, v_j)$ denote the graph distance between two nodes $v_i, v_j \in V$ based on their hierarchical relations. Also, let $d_{\mathbb{B}}(p_i, p_j)$ denote the hyperbolic distance between two points $p_i,p_j \in \mathbb{B}^d$ in the Poincar\'e ball. We seek to obtain a set of prototypes $P=\{p_i\}_{i=1}^{|V|}$ corresponding to $V=\{v_i\}_{i=1}^{|V|}$ such that $d_{\mathbb{B}}(p_i, p_j) \propto d_{\mathcal{T}}(v_i, v_j)$, \emph{i.e.,} the hyperbolic distance $d_{\mathbb{B}}$ is proportional to the edge distances in graph $\mathcal{T}$. The hyperbolic distance $d_\mathbb{B}$ between two points $p_1,p_2 \in \mathbb{B}^d$ is given by the following equation:
\begin{equation}
        d_{\mathbb{B}}({p_1}, {p_2}) =  \frac{2}{\sqrt{c}} \arctanh (\sqrt{c} ||-p_1 \oplus_c p_2||).
\end{equation}
where $\langle p_1, p_2 \rangle$ denotes the inner product of two vectors $p_1$ and $p_2$ and $\oplus$ denotes the M\"{o}bius addition given by:
\begin{equation}
    p_1 \oplus_c p_2 = \frac{(1 + 2c \langle p_1, p_2 \rangle + c||p_2||^2)p_1 + (1 - c||p_1||^2)p_2}{1 + 2c \langle p_1, p_2 \rangle + c^2||p_1||^2 ||p_2||^2}
\end{equation}
We minimize the loss function $\mathcal{L}_{Poincar\acute{e}}$ defined by \cite{poincare}. Let $\mathcal{R}=\{(u,v)\}$ denote the set of transitive closures in graph $\mathcal{T}$, meaning there is a path from node $u$ to $v$. \cite{entailment} define $\mathcal{R}$ as entailment relations, where $v$ entails $u$, or equivalently, that $u$ is a subconcept of $v$. $\mathcal{L}_{Poincar\acute{e}}$ encourages semantically similar objects to be close in the embedding space according to their Poincar\'e distance:
\begin{equation}
\end{equation}

$\mathcal{N}(u)=\{v| (u, v) \notin \mathcal{R}\}\cup \{u\} $ denotes the set of negative examples for $u$. Negative examples $\mathcal{N}(u)$ include all nodes $v$ that do not entail $u$.
While the Poincar\'e loss results in hyperbolic prototypes $P$ in tree-shaped regions on $\mathbb{B}^d$, there are no guarantees of entailment, \emph{i.e.,} of a partial order relationship that requires the region of each subtree to be fully covered by their parent tree. Therefore, following \cite{entailment}, we apply a max-margin entailment loss $\mathcal{L}_{Entailment}$ on the extracted Poincar\'e embeddings $P$, to enforce entailment regions:

\begin{equation}
\label{eq:entailment}
    \mathcal{L}_{Entailment} = \sum_{(u, v) \in \mathcal{R}} E(u,v) + \sum_{(u',v')\in \mathcal{N}} max(0,\gamma-E(u',v')).
\end{equation}
The energy function $E(u, v):= \max (0, \Xi(u,v) - \psi(v))$ measures how far point $u$ from belonging to the entailment cone $\psi(v)$ is. The first term encourages $u$ to be part of the entailment cone $\psi(v)$ for $(u,v)\in\mathcal{R}$. The second term pushes negative samples $(u',v')\in \mathcal{N}$ angularly away for a minimum margin $\gamma >0$ if they don’t share an entailment cone. For full details of Equation \ref{eq:entailment}, we refer to \cite{entailment}.

We observe that the entailment loss tends to bring the prototypes, especially those of the instances $V_I$, too close to each other. To counteract this, we apply a separation loss $\mathcal{L}_S$, similar to the approach in \citet{actionsonhyperbole}, which ensures that all prototypes $P$ are adequately separated.
\vspace{-2mm}
\begin{equation}
    \mathcal{L}_S(P) = \vec{1}^T(\Bar{P} \Bar{P}^T - I)\vec{1},
\end{equation}
where $\Bar{P}$ denotes the vector-wise $l_2$-normalization of $P$. The proposed loss function minimizes the cosine similarity between any two prototypes. \reb{Algorithm \ref{alg:first_stage} summarizes the first stage of HyperCLIC, detailing the sequential application of the three losses that result in the hyperbolic prototypes.} After obtaining the prototypes for all nodes in the tree $\mathcal{T}$, we only use the prototypes $P=\{P_y\}_{y=1}^{|V_I|}$ corresponding to the instance (leaf) nodes $V=\{v_y\}_{y=1}^{|V_I|}$ below.

\subsection{Continual hyperbolic learning}
Our proposed hyperbolic continual learner draws inspiration from the core design choices of iCaRL \citep{icarl}, the best-performing model for instance-level classification on the EgoObjects dataset according to \cite{challenge_winners}.
We use a backbone $\varphi(.;\theta_t)$ to extract features from a training sample $x_i$, where $\theta_t$ denotes the model parameters at timestep $t$. The extracted features form a Euclidean representation $\varphi(x_i;\theta_t)$. We assume that the backbone output is in the tangent space $T_x \mathcal{M}$, while the extracted prototypes from Section \ref{sec:formalize_hierarchy} are in the hyperbolic space $\mathcal{M}$. Thus, we need to project the Euclidean representation to the hyperbolic space $\mathcal{M}$ through an exponential map, which embeds visual representation into the hyperbolic space where the hierarchy of classes and instances are embedded as prototypes:
\vspace{-4mm}
\begin{align}
    \mathrm{z}_i^t &= \exp_{0}{(\varphi(\mathrm{x_i};\theta^t))}, & \text{where: } \exp_0^c{(\mathrm{x})} &= \tanh{(\sqrt{c} \|\mathrm{x}\|)} \cdot \frac{\mathrm{x}}{\sqrt{c}\|\mathrm{x}\|}.
\end{align}

The exponential map definition is borrowed from \citep{ganea2018hyperbolic}, where $c$ denotes the curvature of the hyperbolic space. A higher $c$ indicates a more curved space. Here, $\mathrm{z_i}^t$ is the hyperbolic representation of $\varphi(x_i;\theta_t)$. Our goal is to minimize the hyperbolic distance of representation $\mathrm{z}$ with its instance prototype $P_y$. Thus, we define the hyperbolic logit $h(\mathrm{z},y)$ as the negative hyperbolic distance of vector representation $\mathrm{z}$ and all prototypes $y$. The conditional probability $p(y_i|\mathrm{z}_i^t)$ can be derived by the softmax of the hyperbolic logit:
\vspace{-2mm}
\begin{align}
      p(y_i|\mathrm{z}_i^t) &= \frac{e^{h(z_i^t,y_i)}}{\sum_{j=1}^{|C|} e^{h(z_i^t,y_j)}}, & \text{where: }  h(\mathrm{z},y) &= -\frac{d_{\mathbb{B}}(\mathrm{z}, P_y)}{\tau}.
\end{align}
The temperature parameter $\tau$ controls the entropy of the probability distribution while preserving the relative ranks of each class. We can make the probability distribution more peaked or smoother by adjusting this parameter. \\Once the hyperbolic logits are calculated, we can compute the classification loss as the cross entropy between the hyperbolic logits and the targets:
\vspace{-4mm}
\begin{equation}
        \Lagr_{cls} = -\frac{1}{|\mathcal{D}^t|} \sum_{(x_i,y_i)\in \mathcal{D}^t}\sum_{i=1}^{|C|}y_i \cdot \log (\frac{e^{h(z_i^t,y_i)}}{\sum_{j=1}^{|C|} e^{h(z_i^t,y_j)}}).
\end{equation}

While the classification loss ensures that the new classes learn their representation, the ultimate goal of class-incremental learning is to continually learn a model that works both for the old and the new classes. Formally, the model should not only acquire the knowledge from the current task $\mathcal{D}^t$ but also preserve the knowledge from former tasks $\mathcal{D}^{b<t}$. We use a hyperbolic distillation loss to ensure that the model predicts the same hyperbolic logits for the old classes $\mathcal{D}^{b<t}$ at time $t$ as time $t-1$. Following \cite{icarl}, we keep a copy of the model parameters at time $t-1$ as $\theta^{t-1}$ and define the probability distribution at time $t-1$ as $p(y_i|\mathrm{z}_i^{t-1})$ where $(x_i, y_i) \in \mathcal{D}^{b<t}$:
\vspace{-3mm}
\begin{equation}
     p(y_i|\mathrm{z}_i^{t-1}) = \frac{e^{h(z_i^{t-1},y_i)}}{\sum_{j=1}^{|C|} e^{h(z_i^{t-1},y_j)}}.
\end{equation}
The hyperbolic distillation loss is calculated as the cross-entropy loss between the probability distribution at time $t$ and $t-1$:
\vspace{-3mm}
\begin{equation}
        \Lagr_{distil} = -\frac{1}{|\mathcal{D}^{b<t}|} \sum_{(x_i,y_i)\in \mathcal{D}^{b<t}}\sum_{i=1}^{|C^{b<t}|}(\frac{e^{h(z_i^{t-1},y_i)}}{\sum_{j=1}^{|C^{b<t}|} e^{h(z_i^{t-1},y_j)}}) \cdot \log (\frac{e^{h(z_i^t,y_i)}}{\sum_{j=1}^{|C^{b<t}|} e^{h(z_i^t,y_j)}}).
\end{equation}

\reb{The final loss of the second stage is $\Lagr = \Lagr_{cls}+\Lagr_{distil}$ following \cite{icarl}. Algorithm \ref{alg:second_stage} summarizes the second stage of HyperCLIC, where the hyperbolic prototypes from the first stage are used to classify new classes at the current task and distill hyperbolic logits from previousely seen classes.} Samples from old classes $\mathcal{D}^{b<t}$ are selected at the end of each task by the herding strategy \citep{welling2009herding}, which is a commonly used strategy aiming to select the most representative samples of each class. These exemplars are saved in memory and are representative data points of each known instance class. In addition to distillation loss, the exemplars are also used during inference. During inference, the closeness to exemplars determines the final prediction for a given data point. We perform the nearest mean of exemplar classification following \cite{icarl}: $
    y_i^*=\argmin_{y=1,...,n_I}{\|\varphi(\mathrm{x_i})-\mu_y\|}$, where $\mu_y$ denotes the mean of exemplars for the instance class $y$ and $n_I$ denotes the number of instance classes. The exemplars and their means could be calculated in hyperbolic space, however, we observe that the Euclidean representations are already aligned with the hierarchy and conclude that there is no need for an extra hyperbolic computation.
Overall, our method aligns classification and distillation losses with hyperbolic representations of a fixed hierarchy, leveraging the class-instance hierarchy with minor modifications to existing continual learning methods. This alignment allows continual classification at multiple levels of granularity such as instance and class-level.

\begin{table}[]
\centering
\begingroup 
\setlength{\tabcolsep}{2.5pt}
\caption{\textbf{Experimental results on EgoObjects} with a ResNet34 backbone. HyperCLIC outperforms all methods with and without pretraining. Methods marked with $\sharp$ are significantly tested against \mymethod\ at $p < 0.05$ (See \ref{sec:statisticall_significance}).}
\label{tab:SOTA:egoobject:all}
\begin{tabular}{lcccccccccc}
\toprule
& \multicolumn{5}{c}{\textbf{From scratch}} & \multicolumn{5}{c}{\textbf{Pretrained}} \\
\cmidrule(lr){2-6} \cmidrule(lr){7-11}
  & \multicolumn{3}{c}{\textbf{Accuracy $\uparrow$}} & \textbf{LCA $\downarrow$} & \textbf{Forgetting $\downarrow$} & \multicolumn{3}{c}{\textbf{Accuracy $\uparrow$}} & \textbf{LCA $\downarrow$} & \textbf{Forgetting $\downarrow$}\\
                       & Instance & Class & Superclass & & & Instance & Class & Superclass & & \\ \midrule
Naive & 0.44 & 1.07 & 1.58 & 5.74 & 10.38 & 6.06 & 15.55 & 18.24 & 4.70 & 93.03 \\
EWC  &0.22& 0.66 & 1.42 & 5.26 & 10.38 &  6.07& 15.77 & 18.37 & 4.74 & 92.83\\
iCaRL $\sharp$  & 20.05 & 21.39 & 22.24 & 5.45 & 10.57 &  81.63 & 87.02 & 87.81 & 3.89 & \textbf{3.77}\\
CoPE & 3.14 & 4.45 & 4.91 & 5.63 & 25.38 & 37.15 & 49.89 & 52.51 & 4.22 & 32.40\\
GDumb & 0.50& 1.27& 1.88&5.50 & 5.87&  61.51&71.00 & 72.82& 4.02& 18.22\\
DER $\sharp$ &19.59 & 20.59& 21.14& 5.61&35.39 &  80.00& 84.40& 85.20&4.28 &10.59\\
\rowcolor{Gray} \mymethod $\sharp$ & \textbf{41.76} & \textbf{45.91} & \textbf{48.04} & \textbf{4.93} & \textbf{4.17} &  \textbf{84.81} & \textbf{91.94} & \textbf{92.67} & \textbf{2.99} & 7.08\\ 
\bottomrule
\end{tabular}
\endgroup
\end{table}


\section{Results}

\paragraph{\rebuttal{Dataset}}
\rebuttal{ We focus on EgoObjects \citep{egoobjects} since it is the only large-scale class-incremental dataset with a significant number of instances \textit{and} classes and a non-trivial hierarchy, reflecting real-world complex hierarchical structures. Core50 dataset \citep{core50} is also an instance-level dataset, but is a toy dataset with only 10 classes with 5 instances each, resulting in minimal differences between methods. We see a great potential for additional large instance-level continual datasets due to the real-world relevance, but there is unfortunately a lack of large-scale instance-level continual datasets.
The EgoObjects benhcmark~\citep{egoobjects}, first introduced at the 3rd CLVISION Challenge~\citep{clvision_challenge}, represents a challenging benchmark for continual learning, particularly because of the instance-level object understanding with $1110$ unique instances across $277$ classes. The objects follow an imbalanced long-tailed distribution with varying clutter, occlusion, lighting, and backgrounds. The order of classes follows scenarios of ever-evolving environments with the reoccurance of previously seen objects \citep{hemati2025continual}. 
This dataset comprises a stream of 15 tasks, each created by cropping the main object from egocentric videos with each task including 74 instances. Since the test set labels are not provided in the challenge, we split the training set into training ($80$\%), testing ($10$\%), and validation ($10$\%) subsets. Following the challenge convention, we divide the dataset into non-overlapping sets of \texttt{gaia\_ids}, which are identifiers for each unique video clip. For reproducibility purposes, we have provided the code and metadata including the hierarchy and the train, test and validation sets at \href{https://anonymous.4open.science/r/HyperCLIC/}{this} link.}

\rebuttal{\paragraph{Hierarchies}
For the EgoObjects~\citep{egoobjects} dataset, we construct a hierarchy using WordNet~\citep{wordnet}, a comprehensive lexical database of the English language. We match each class in the dataset with the most frequent noun synset in WordNet. To find the path from the leaf nodes to the root node, we iteratively choose the hypernym with the shortest path to the root. If this path does not include the \texttt{physical\_entity.n.01} node, we select the next shortest path that does. We manually add classes that do not match any synsets to the hierarchy. Categories without any instances ($91$ out of $277$) are removed. We also remove nodes with only one child, except for parents of leaf nodes. The resulting unbalanced hierarchy has a maximum of $7$ instances per category, a minimum of $1$, and a depth of $12$. Our automated hierarchy extraction method can be adapted to other datasets. The hierarchy is provided with the code at \href{https://anonymous.4open.science/r/HyperCLIC/}{this} link to enable reproducibility.}

\rebuttal{\paragraph{Implementation details}
For \mymethod EgoObjects experiments, we train a ResNet34 backbone for $3$ epochs following \cite{challenge_winners}, with a batch\_size of $90$, and a learning rate of $0.01$ for $15$ tasks. The memory size is $3500$ according to the challenge. The pretrained models are pretrained with \texttt{ResNet34\_Weights.IMAGENET1K\_V1}, \texttt{Wide\_ResNet50\_2\_Weights.IMAGENET1K\_V2} and \texttt{RegNet\_X\_16GF\_Weights.IMAGENET1K\_V2}. In all datasets, the curvature of the Poincar\'e model (\texttt{Poincar\'eBallExact} \cite{geoopt}) is $1$. The hyperbolic prototypes are $64$-dimensional using \cite{entailment} with $150$ Poincar\'e epochs and $50$ entailment epochs, plus separation following \cite{actionsonhyperbole} with a learning rate of $1.0$ for $500$ epochs. } \rebuttal{For all experiments, we choose the hyperparameters of our method with minimal hyperparameter tuning for fair comparison. Following hyperbolic learning literature, we set the temperature parameter $\tau$ to a common value of $0.1$ \citep{ibrahimi2024intriguing, actionsonhyperbole,sinha2024learning}. Parameter $\lambda$ was fixed at $0.5$ to balance distillation and classification objectives equally. We show that the results can be improved with better backbones. For consistency, we applied the same hyperparameters (optimized for 15-task scenarios) across all evaluations. As shown in Table \ref{tab:SOTA:egoobject:6_and_37_tasks}, this leads to some performance variations - for instance, DER achieves better results than HyperCLIC on the 6-task scenario. All methods would probably benefit from extensive tuning but that was not explored in the paper. }

\paragraph{Baselines} We evaluate HyperCLIC against four canonical CIL methods representing key strategies: EWC \citep{ewc} (regularization-based), iCaRL \citep{icarl} and GDumb \citep{prabhu2020gdumb} (replay-based), and DER \citep{der} (architecture expansion). Additionally, we compare to CoPE \citep{cope}, an online prototype-based approach well-suited for EgoObjects, and naive fine-tuning. \rebuttal{The baseline methods were designed without hierarchical information, and there is no non-trivial way to integrate hierarchical structure into their existing frameworks. This underlines the importance of our proposed method and the fact that other continual learning methods could benefit from including the hierarchical information in future research. Comparison to \cite{gao2023exploring} was unfortunately not possible due to the unavailability of their code. Our goal is to advocate joint instance- and class-level continual learning and hyperbolic embeddings' potential to enrich hierarchical representations in standard continual learning scenarios. We therefore focus on the most well-known approaches in continual learning. We believe that generalizing recent instance-level approaches to our task is an interesting future research direction.} Our implementations are based on \cite{avalanche}. ICaRL is particularly relevant as it is the best-performing model for instance-level classification on EgoObjects \citep{challenge_winners}. 
\paragraph{Comparison to existing methods}
Table \ref{tab:SOTA:egoobject:all} reports the results of different baselines on the EgoObjects dataset with 15 tasks. If trained from scratch, \mymethod achieves a significantly higher instance-level accuracy ($41.76\%$), class-level accuracy ($45.91\%$), and superclass-level accuracy ($48.04\%$) compared to all baselines. The closest competitor is iCaRL \citep{icarl} with an instance-, class-, and superclass-level accuracy of $20.05\%$, $21.39\%$, and $22.24\%$, respectively, which is less than half of \mymethod's performance. This demonstrates HyperCLIC's strength in maintaining performance across different levels of granularity. \mymethod shows a lower LCA value ($4.93$) indicating that the inaccurate predictions are less hierarchically severe compared to other baselines. The second-best is EWC \citep{ewc} with $5.26$. \mymethod also shows the lowest forgetting rate ($4.17\%$), indicating better retention of learned knowledge. \rebuttal{The low scores of baselines is due to the large number of instances and the long-tailed distribution. They fail to differentiate between instances in the EgoObjects dataset even in one task, because of the challenging nature of the dataset. In general, instance-level continual learning faces different challenges compared with class-level continual learning.} For the pretrained scenario, \mymethod leads with the instance-level accuracy of $84.81\%$, followed by iCaRL \citep{icarl} with $81.63\%$, and DER \citep{der} with $80\%$. DER is a strong non-transformer-based continual baseline. Across both trained-from-scratch and pretrained scenarios, \mymethod consistently outperforms all baselines on every metric. As shown in Table \ref{tab:SOTA:egoobject:6_and_37_tasks}, \mymethod achieves the highest performance in the challenging 37-task setting and ranks second in instance- and class-level accuracy for 6 tasks. \rebuttal{The performance differences between DER, HyperCLIC, and iCaRL on EgoObjects in Table \ref{tab:SOTA:egoobject:6_and_37_tasks} can be attributed to their distinct approaches to continual learning. DER demonstrates stronger performance in the 6-task scenario due to its inherent robustness with fewer tasks, though this advantage diminishes with the 37-task setup where its performance declines more significantly. HyperCLIC maintains more stable results across different task quantities thanks to its hierarchical structure, which provides consistent representation learning despite being slightly outperformed by DER in the 6-task case. Meanwhile, iCaRL shows greater sensitivity to task quantity variations, with performance dropping for both 6 and 37 tasks compared to the 15-task baseline, though its smaller performance gap between 6 and 37 tasks suggests potential for improvement through hyperparameter optimization. All methods in Table \ref{tab:SOTA:egoobject:6_and_37_tasks} used identical hyperparameters as Table \ref{tab:SOTA:egoobject:all}, indicating that task-specific tuning could further enhance their results.} HyperCLIC's excelling in instance-, class-, and superclass-level accuracy while maintaining lower LCA and minimal forgetting, \mymethod demonstrates its ability to learn hierarchically-aware representations while preserving knowledge from previous tasks.

Given that iCaRL \citep{icarl} has the highest performance after HyperCLIC, we use it as the primary comparison for evaluating our method in subsequent experiments. Table \ref{tab:SOTA:egoobject:per_experience} illustrates the instance-level accuracy for each task (T1 to T15) after the completion of all tasks. In both non-pretrained and pretrained models, iCaRL \citep{icarl} demonstrates higher stability but at the cost of not learning effective representations for new tasks, thereby compromising its performance on recent tasks to retain knowledge from previous ones. We can also observe this behavior in Table \ref{tab:SOTA:egoobject:all}. In contrast, our method not only preserves knowledge from prior tasks but also achieves higher accuracy on the latest tasks compared to the baseline.

\begin{table}[]
\centering
\begingroup 
\setlength{\tabcolsep}{9pt}
\caption{\textbf{Experimental results on EgoObjects with 6 and 37 tasks} from scratch with a ResNet34 backbone.}
\label{tab:SOTA:egoobject:6_and_37_tasks}
\begin{tabular}{lcccccccccc}
\toprule
                       & \multicolumn{2}{c}{\textbf{Instance $\uparrow$}} & \multicolumn{2}{c}{\textbf{Class $\uparrow$}} & \multicolumn{2}{c}{\textbf{Superclass $\uparrow$}} & \multicolumn{2}{c}{\textbf{LCA $\downarrow$}} & \multicolumn{2}{c}{\textbf{Forgetting $\downarrow$}} \\ 
                       \cmidrule(lr){2-3} \cmidrule(lr){4-5}
                       \cmidrule(lr){6-7}
                       \cmidrule(lr){8-9}
                       \cmidrule(lr){10-11}
                       & 6 & 37 & 6 & 37 & 6 & 37 & 6 & 37 & 6 & 37 \\ 
\midrule
Naive &  02.53 & 00.21 & 03.48 & 1.13 & 04.11 & 1.74 & 5.73 & 5.54 & 22.10 & 10.60
\\
EWC  & 02.47 & 00.25 & 03.51 & 1.31 & 04.52 & 1.31 & 5.62 & 7.69 & 21.99 & 11.52
\\
iCaRL  &  23.56 & \underline{17.46} & 24.74 & \underline{18.74} & 25.69 & \underline{19.57} & \underline{5.41} & 5.48 & 6.60 & 12.22
\\
GDumb & 00.56 & 00.71 & 01.24 & 1.75 & 01.36 & 2.39 & 5.59 & \underline{5.30} & \underline{5.37} & \underline{6.46}
\\
DER &  \textbf{36.01} & 2.05 & \textbf{36.99} & 2.85 & \underline{37.75} & 3.34 & 5.42 & 5.74 & 12.70 & 45.76
\\
\rowcolor{Gray}\mymethod &  \underline{34.09} & \textbf{39.86} & \underline{36.71} & \textbf{47.18} & \textbf{38.06} & \textbf{49.81} & \textbf{5.12} & \textbf{4.48} & \textbf{00.26} & \textbf{5.73}
\\ 
\bottomrule
\end{tabular}
\endgroup
\end{table} 
\vspace*{-5mm}
\begin{table*}[t]
\centering
\begingroup 
\setlength{\tabcolsep}{2pt}
\begin{minipage}{0.49\textwidth}
\centering
\caption{\textbf{Balancing the effect of classification and distillation losses in EgoObjects dataset.} Equally weighting the losses yields the highest performance in HyperCLIC.}
\label{tab:lambda}
\color{black}
\begin{tabular}{lccccc}
\toprule
 \textbf{$\lambda$} & \multicolumn{3}{c}{\textbf{Accuracy $\uparrow$}} & \textbf{LCA $\downarrow$} & \textbf{Forgetting $\downarrow$}\\
                        & Instance & Class & Superclass & & \\ \midrule
0.1    & 35.33 & 39.71 & 42.09 & 4.90 & 7.28       \\
0.3    & 37.54 & 42.05 & 44.51 & 4.83 & 6.18       \\
\rowcolor{Gray} Ours   & 41.76 & 45.91 & 48.04 & 4.93   & 4.17       \\
0.7    & 38.02 & 41.56 & 43.38 & 4.92 & -0.59      \\
0.9    & 24.49 & 26.55 & 27.83 & 5.32 & -4.92      \\                    
\bottomrule
\end{tabular}
\end{minipage}
\vspace*{-5mm}
\endgroup
\hfill
\begin{minipage}{0.48\textwidth}
\centering
\includegraphics[width=\textwidth, trim={32cm 4cm 32cm 3cm}, clip]{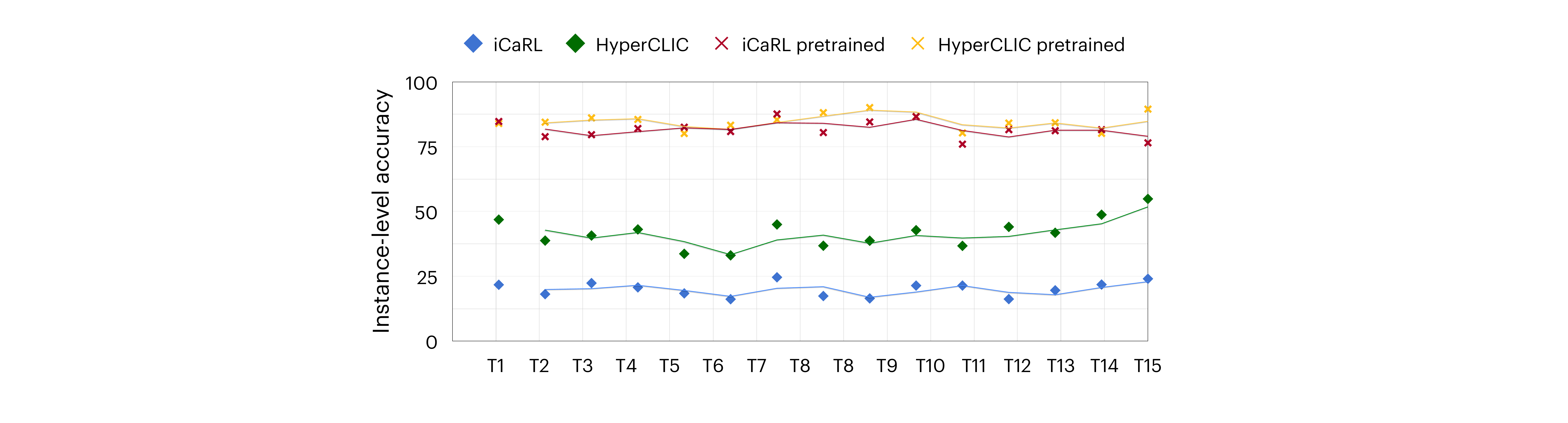}
\caption{\textbf{The final per-task accuracies and moving average on EgoObjects}. HyperCLIC outperforms iCaRL by maintaining prior knowledge while delivering higher accuracy on new tasks.}
\vspace*{-2mm}
\label{tab:SOTA:egoobject:per_experience}
\end{minipage}
\end{table*}

\begin{wrapfigure}{r}{0.7\textwidth}
\centering
\includegraphics[width=\linewidth, trim={37cm 7cm 40cm 7cm}, clip]{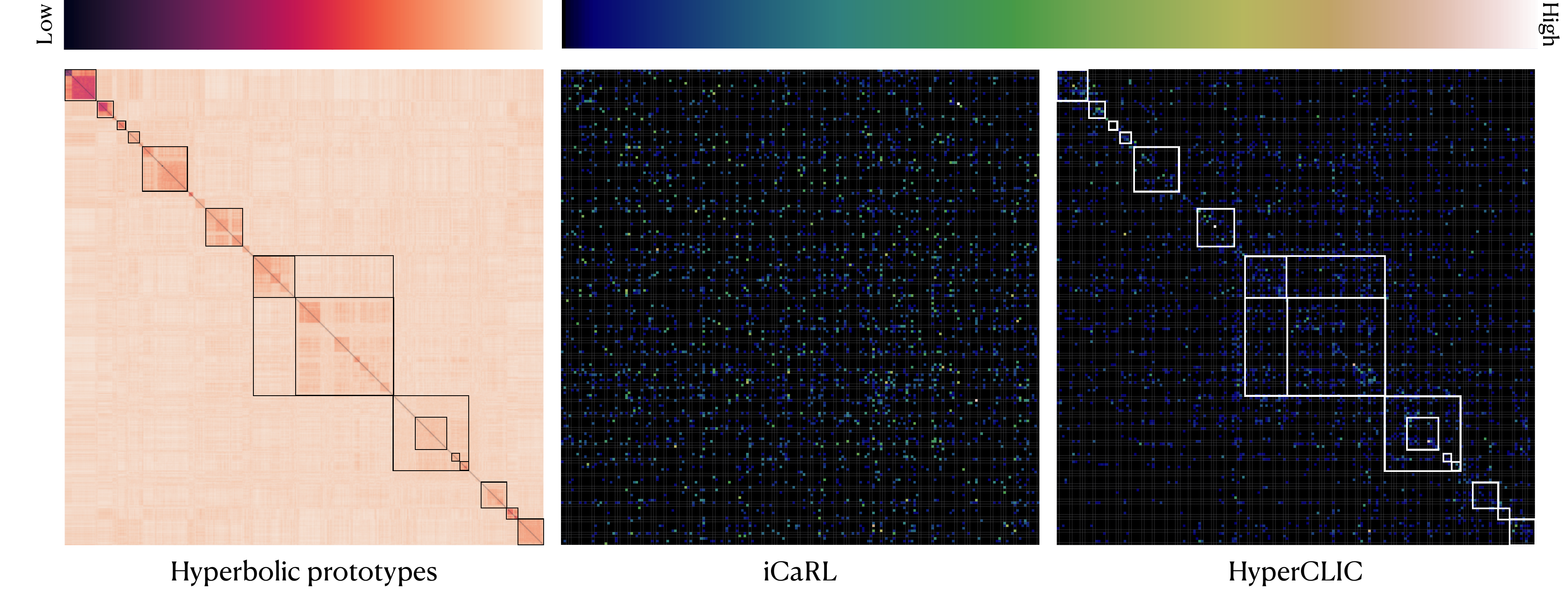}
\caption{\textbf{Left:} the pair-wise distances of the instance-level hyperbolic prototypes. \textbf{Middle \& right:} the class-level predictions only for the wrong instance-level predictions. The squares highlight the hierarchical structure of \mymethod mistakes compared with iCaRL and its similarity to the original hierarchy.}
\label{fig:confusion_class_level}
\vspace*{-5mm}
\end{wrapfigure}

\paragraph{Class-level errors} Figure \ref{fig:confusion_class_level} compares the class-level confusion matrix for incorrectly predicted instances between iCaRL and HyperCLIC, alongside the distances of hyperbolic prototypes. In this matrix, the squares represent children of the same parent node, and larger squares encompass multiple subtrees. The aim of this experiment is to demonstrate that even when instances are misclassified, HyperCLIC's class-level predictions still follow the hierarchical structure. As shown, iCaRL's class-level predictions are evenly distributed, while HyperCLIC's predictions are primarily concentrated along the diagonal. Additionally, a similar hierarchical organization is observed in \mymethod when compared with the pairwise hyperbolic prototype distances, represented by matching squares. This indicates that \mymethod maintains the hierarchical structure, even when making incorrect predictions, resulting in less severe errors. \rebuttal{Figure \ref{fig:confusion_class_level} is the qualitative demonstration of the quantitative LCA metric. This figure shows the class-level predictions only for the wrong instance-level predictions and highlights the hierarchical structure of the mistakes of HyperCLIC. LCA measures the number of edges between the incorrect prediction and the lowest common ancestor of the ground-truth and prediction. Intuitively, if an object is incorrectly predicted at instance level but correctly at class level, the LCA is equal to 1. If an object is incorrectly predicted at instance level but correctly at superclass level, the LCA is equal to 2 instead.}
\vspace*{-1mm}

\begin{figure}
\centering
\includegraphics[width=\linewidth, trim={20cm 2cm 20cm 0cm}, clip]{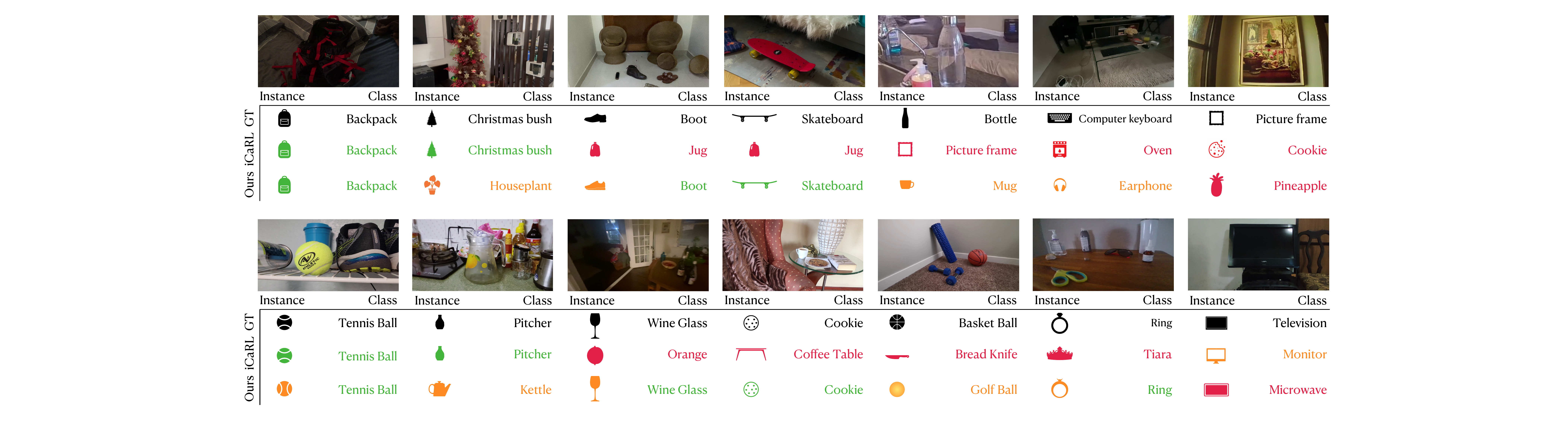}
\caption{\textbf{Qualitative examples from EgoObjects} comparing instance and class-level predictions for ground truth (1\textsuperscript{st} row), iCaRL (2\textsuperscript{nd} row), and our method (3\textsuperscript{rd} row). \textcolor{green}{Green} indicates correct predictions, while \textcolor{red}{red} and \textcolor{orange}{orange} denote severe and less severe hierarchical errors, respectively. \mymethod makes less severe errors compared to iCaRL.}
\label{fig:qualitative_examples}
\end{figure}

\paragraph{Qualitative examples: success and failure cases} Figure \ref{fig:qualitative_examples} presents several images from the test set and compares the instance and class-level predictions of our method against the baseline. \rebuttal{In the first row and the first column, both methods easily make correct instance-level predictions. Whereas, in the far-right column in the first row, it is challenging for both methods, as it depicts food items in a drawing, leading both to predict unrelated instances and classes. The middle examples reveal that when \mymethod makes an incorrect instance-level prediction, the class-level prediction remains hierarchically close. \mymethod makes more instance-level mistakes, while correctly classifying at class level. Another distinction of \mymethod is that class-level mistakes are hierarchically correct. For example, it might mistake a ``Christmas bush'' for a ``house plant'' or a ``bottle'' for a ``mug'', whereas, iCaRL mistakes ``basket ball'' for a ``bread knife'' or a ``wine glass'' for an ``orange''.} This suggests that by incorporating hierarchical information during training, \mymethod reduces the severity of hierarchical errors.
\paragraph{\reb{Temperature}} \reb{In the second stage of HyperCLIC, one of the key hyperparameters that significantly impacts the continual learning of new classes in the classification loss is the temperature. The temperature, $\tau$, controls the smoothness of the probability distribution derived from the pairwise distances of the hyperbolic prototypes. In hyperbolic literature \citep{ibrahimiintriguing, actionsonhyperbole}, $\tau$ is commonly set to $0.1$. Our results, as shown in Table \ref{tab:temperature}, indicate that \mymethod is highly sensitive to the choice of temperature, especially when trained from scratch compared to using a pretrained model. The best performance is consistently achieved with a temperature of $0.1$ in both scenarios. When the probability distribution is either too peaked or too smooth, the model struggles to effectively distinguish between classes.}
\begin{table*}[]
\centering
\begingroup 
\setlength{\tabcolsep}{4pt}
\caption{\reb{\textbf{The effect of different temperatures on \mymethod in EgoObjects.} Our setting with temperature of 0.1 is highlighted in gray. Our method is more sensitive to the temperature in the from-scratch setting.}}
\label{tab:temperature}
\color{black}
\begin{tabular}{lccccc|ccccc}
\toprule
& \multicolumn{5}{c}{\textbf{From scratch}} & \multicolumn{5}{c}{\textbf{Pretrained}} \\
\cmidrule(lr){2-6} \cmidrule(lr){7-11}
& Instance & Class & Superclass & LCA & Forgetting & Instance & Class & Superclass & LCA & Forgetting \\
\midrule

\reb{0.01}                        & \reb{26.14} & \reb{28.26} & \reb{29.18} & \reb{5.38} & \reb{-1.81}      & \reb{77.79} & \reb{84.90} & \reb{85.64} & \reb{3.92} & \reb{8.06}       \\
\reb{0.05}                        & \reb{11.79} & \reb{12.93} & \reb{13.78} & \reb{5.54} & \reb{4.52}       & \reb{82.21} & \reb{90.02} & \reb{91.16} & \reb{2.97} & \reb{6.67}       \\
\reb{0.09}                        & \reb{38.21} & \reb{42.59} & \reb{44.44} & \reb{4.83} & \reb{3.54}       & \reb{82.03} & \reb{89.93} & \reb{91.13} & \reb{2.91} & \reb{5.21}       \\
\rowcolor{Gray} \reb{Ours}                   & \reb{41.76} & \reb{45.91} & \reb{48.04} & \reb{4.93}   & \reb{4.17}       & \reb{84.81} & \reb{91.94} & \reb{92.67} & \reb{2.99}   & \reb{7.08}       \\
\reb{0.3}                         & \reb{19.76} & \reb{24.07} & \reb{27.22} & \reb{4.94} & \reb{11.82}      & \reb{79.22} & \reb{88.76} & \reb{89.88} & \reb{2.88} & \reb{7.70}       \\
\reb{0.5}                         & \reb{16.35} & \reb{20.77} & \reb{23.68} & \reb{4.97} & \reb{9.11}       & \reb{76.80} & \reb{87.35} & \reb{88.72} & \reb{2.74} & \reb{6.34}       \\
\reb{1.0}                         & \reb{12.67} & \reb{16.52} & \reb{19.21} & \reb{5.03} & \reb{10.39}      & \reb{71.86} & \reb{85.35} & \reb{87.21} & \reb{2.62} & \reb{6.47}       \\

\bottomrule
\end{tabular}
\endgroup
\end{table*}
\paragraph{Distillation loss} A key component of our method is the distillation loss, which ensures that the model retains hyperbolic knowledge from previous tasks. To evaluate this, we tested three different loss functions: Kullback-Leibner (KL) divergence, Mean Squared Error (MSE), and cross-entropy loss. Table \ref{tab:ablation:distillation:egoobject} presents the results for both pretrained and from-scratch scenarios. KL divergence shows the lowest forgetting rates, at $-1.40$\% and $-6$\% respectively, for from-scratch and pretrained scenarios. However, it performs the worst in terms of hierarchical metrics and instance accuracy, suggesting it disrupts the balance between plasticity and stability by favoring stability at the expense of plasticity. Cross-entropy loss performs the best across both scenarios, which is why our method is based on this loss function. There is also a more notable difference between MSE and cross-entropy when training from scratch compared to the pretrained scenario.


\begin{table*}[]
\centering
\begingroup 
\setlength{\tabcolsep}{2.5pt}
\caption{\textbf{EgoObjects results with different distillation losses} with a ResNet34 backbone. Cross-entropy loss consistently outperforms other distillation losses in both scenarios.}
\label{tab:ablation:distillation:egoobject}
\begin{tabular}{lccccc|ccccc}
\toprule
  & \multicolumn{5}{c}{\textbf{From scratch}} & \multicolumn{5}{c}{\textbf{Pretrained}}\\
  \cmidrule(lr){2-6} \cmidrule(lr){7-11}
                       & Instance & Class & Superclass & LCA & Forgetting & Instance & Class & Superclass & LCA & Forgetting\\ \midrule
KL divergence & 32.60 & 35.30 & 36.95 & 5.13 & -1.40 & 49.76 & 62.45 & 65.46 & 3.75 & -6.00 \\ 
MSE & 36.09 & 40.33 & 42.53 & 4.94 & 8.18 &  84.38 & 91.21 & 92.33 & 2.92 & 7.84 \\
\rowcolor{Gray} Cross-entropy &   41.76 & 45.91 & 48.04 & 4.93 & 4.17 & 84.11 & 91.61 & 92.54 & 2.96 & 7.08 \\
\bottomrule
\end{tabular}
\endgroup
\end{table*}



\paragraph{Balancing classification and distillation losses} The continual learning objective in the second stage of \mymethod combines classification and distillation losses, following iCaRL. To evaluate the impact of each loss term, we use a balancing factor, $\lambda$, as shown in the loss function: $\Lagr = \lambda \cdot \Lagr_{distil} + (1-\lambda) \cdot \Lagr_{cls}$. By varying $\lambda$, we assess the influence of classification and distillation losses on the model's performance. Table \ref{tab:lambda} presents the results for different $\lambda$ values. Our findings indicate that a fully balanced loss achieves the best performance across both instance-level accuracy and hierarchical metrics. \rebuttal{The balanced weighting ($\lambda=0.5$) emerges as optimal because it maintains numerical equilibrium between the two losses. Emphasis on distillation ($\lambda>0.5$) compromises learning new tasks, while over-prioritizing classification ($\lambda<0.5$) leads to catastrophic forgetting of previous knowledge. This observation is consistent with iCaRL's formulation where the unweighted sum of classification and distillation losses proves most effective, suggesting that neither objective should dominate during optimization.} This highlights the importance of giving equal weight to both losses to ensure that the model can effectively learn new classes while retaining knowledge of previously learned classes.

\begin{table*}[]
\centering
\begingroup 
\setlength{\tabcolsep}{2.5pt}
\caption{\textbf{Comparison of iCaRL and \mymethod on EgoObjects with various backbones pretrained on ImageNet-1K.} \mymethod consistently outperforms iCaRL regardless of the architecture. This demonstrates that \mymethod is adaptable to any backbone and, when paired with advanced models, enhances joint continual learning of instances and classes.}
\label{tab:backbones:egoobject}
\begin{tabular}{lccccc|ccccc}
\toprule
& \multicolumn{5}{c}{\textbf{iCaRL}} & \multicolumn{5}{c}{\textbf{\mymethod}} \\
\cmidrule(lr){2-6} \cmidrule(lr){7-11}
& Instance & Class & Superclass & LCA & Forgetting & Instance & Class & Superclass & LCA & Forgetting \\
\midrule
WideResNet50 & 82.17 & 87.46 & 88.08 & 4.02 & 3.42 & 86.50 & 94.17 & 94.61 & 2.58 & 6.01 \\
RegNetX\_16GF & 78.63 & 86.64 & 87.58 & 3.55 & 5.10 & 87.22 & 94.11 & 95.01 & 2.59 & 5.99 \\ 
\rowcolor{Gray} ResNet34 & 81.63 & 87.02 & 87.81 & 3.89 & 3.77 & 84.81 & 91.94 & 92.67 & 2.99 & 7.08 \\ 
ViTB/16 & 80.61 & 89.02 & 89.83 & 3.27 & 1.94 & 88.61 & 94.62 & 95.36 & 2.67 & 4.01\\
\bottomrule
\end{tabular}
\endgroup
\end{table*}

\paragraph{Backbones} All results reported thus far utilize the ResNet34 backbone. Table \ref{tab:backbones:egoobject} compares our method with the leading baseline (iCaRL), using various pretrained backbones. The findings indicate that \mymethod consistently surpasses the baseline across the hierarchical metrics when using WideResNet50, RegNetX, ResNet34 and ViTB/16 as the backbone architectures. Notably, \mymethod achieves the highest performance with the ViTB/16 backbone, attaining instance-, class-, and superclass-level accuracies of $88.61$\%, $94.62$\%, and $95.36$\%, respectively, compared to the best iCaRL results for WideResNet, which are $82.17$\%, $87.46$\%, and $88.08$\%. Although \mymethod improves the LCA distance by nearly one edge, leading to better hierarchical mistakes, it shows a higher forgetting rate. These results suggest that our method, particularly when combined with advanced models, has the potential to significantly enhance the joint continual learning of instances and classes. We further analyze the impact of the number of fine-tuned layers in \mymethod compared to Euclidean-based methods (iCaRL) in Appendix \ref{sec:ablation_number_finetuned_layers}.

\section{Conclusion and Discussion}
Instance-level continual learning addresses the challenging task of recognizing and remembering specific instances of object classes in an incremental setup, where new instances appear over time. This approach forms a more fine-grained challenge than conventional continual learning, which typically focuses on incremental discrimination at the class level. In this paper, we introduced a method for continually learning at both instance and class levels, arguing that real-world continual understanding requires recognizing samples at multiple layers of granularity. We observed that classes and instances form a hierarchical structure that can be leveraged to enhance learning at both levels. To this end, we proposed HyperCLIC, a hyperbolic continual algorithm designed for jointly learning instances and classes. We introduced continual hyperbolic classification and hyperbolic distillation, which embed the hierarchical relationships. Our experiments demonstrated that \mymethod operates effectively at multiple levels of granularity. We conducted ablations on different distillation losses, backbone architectures, and pretrained models. Our qualitative analysis provided evidence that \mymethod makes less hierarchically severe mistakes. \mymethod enables real-world continual understanding, where recognizing and remembering both instances and classes over time is crucial.

\rebuttal{Our approach shares conceptual similarities with exemplar-free prototype-based approaches in using compressed representations to mitigate forgetting. The key distinction is that we maintain both exemplars and prototypes to preserve hierarchical relationships—exemplars help anchor fine-grained instance-level features, while hyperbolic prototypes capture the class hierarchy. We believe that the geometric properties of hyperbolic space can benefit exemplar-free approaches in several ways, for instances by including a hierarchical prototype alignment and we see this as an intriguing future research direction.
}

\bibliography{collas2025_conference}
\bibliographystyle{collas2025_conference}

\newpage
\appendix
\section{Appendix}
\label{sec:appendix}
\subsection{HyperCLIC algorithm pseudo code} Our method consists of two stages. The objective of the first stage (Algorithm \ref{alg:first_stage}) is to obtain a set of hyperbolic prototypes, which are later used in the second stage (Algorithm \ref{alg:second_stage}) for the classification loss: 
\begin{minipage}[t]{0.99\textwidth}
\begin{algorithm}[H]
\caption{\\\reb{3.1 Embedding class-instance hierarchies in hyperbolic space}}
\label{alg:first_stage}
\begin{algorithmic}[1]
\REQUIRE \reb{The joint class-instance hierarchy $\mathcal{T}=(V,E)$}
\STATE \reb{We seek to obtain a set of prototypes $P=\{p_i\}_{i=1}^{|V|}$ corresponding to $V=\{v_i\}_{i=1}^{|V|}$ such that $d_{\mathbb{B}}(p_i, p_j) \propto d_{\mathcal{T}}(v_i, v_j)$}
\STATE \reb{// Initialize prototypes with Poincaré loss:}
\FOR{\reb{$e$ in \texttt{\#Poincaré\_Epochs}}}
    \STATE \reb{Minimize loss:  $\mathcal{L}_{Poincar\acute{e}} = \sum \log \frac{e^{-d_B(u, v)}}{\sum_{v' \in \mathcal{N}(u)} e^{-d_B(u, v')}}$}
\ENDFOR
\STATE \reb{// Enforce max-margin entailment regions:}
\FOR{\reb{$e'$ in \texttt{\#Entailment\_Epochs}}}
    \STATE \reb{Minimize loss:  $\mathcal{L}_{Entailment} = \sum E(u,v) + \max(0,\gamma-E(u',v'))$}
\ENDFOR
\STATE \reb{// Enforce separation between prototypes:}
\FOR{\reb{$e''$ in \texttt{\#Separation\_Epochs}}}
    \STATE \reb{Minimize loss:  $\mathcal{L}_S(P) = \vec{1}^T(\Bar{P} \Bar{P}^T - I)\vec{1}$}
\ENDFOR
\STATE \reb{After obtaining the prototypes for all nodes in the tree $\mathcal{T}$, we only use the prototypes $P=\{P_y\}_{y=1}^{|V_I|}$ corresponding to the instance (leaf) nodes $V=\{v_y\}_{y=1}^{|V_I|}$ in the next stage.}
\end{algorithmic}
\end{algorithm}
\end{minipage}
\hfill

\begin{minipage}[t]{0.99\textwidth}
\begin{algorithm}[H]
\caption{\\\reb{3.2 Continual hyperbolic learning with hierarchical prototypes}}
\label{alg:second_stage}
\begin{algorithmic}[1]
\REQUIRE \reb{$P=\{P_y\}_{y=1}^{|V_I|}$}
\STATE \reb{// The classification loss using prototypes from the previous stage:}
\STATE \reb{$\Lagr_{cls} = -\frac{1}{|\mathcal{D}^t|} \sum_{(x_i,y_i)\in \mathcal{D}^t}\sum_{i=1}^{|C|}y_i \cdot \log \left(\frac{e^{h(z_i^t,y_i)}}{\sum_{j=1}^{|C|} e^{h(z_i^t,y_j)}}\right)$}
\STATE \reb{Keep a copy of the model parameters at time $t-1$ as $\theta^{t-1}$ to calculate the distillation loss:}
\STATE \reb{$\Lagr_{distil} = -\frac{1}{|\mathcal{D}^{b<t}|} \sum_{(x_i,y_i)\in \mathcal{D}^{b<t}}\sum_{i=1}^{|C^{b<t}|}p(y_i|\mathrm{z}_i^{t-1}) \cdot \log p(y_i|\mathrm{z}_i^{t})$}
\STATE \reb{Run network training by minimizing $\mathcal{L} = \Lagr_{cls}+\Lagr_{distil}$}
\end{algorithmic}
\end{algorithm}
\end{minipage}

\subsection{Ablation: The effect of the number of fine-tuned layers} 
\label{sec:ablation_number_finetuned_layers}
\begin{figure}{}
  \centering
  \includegraphics[width=\textwidth]{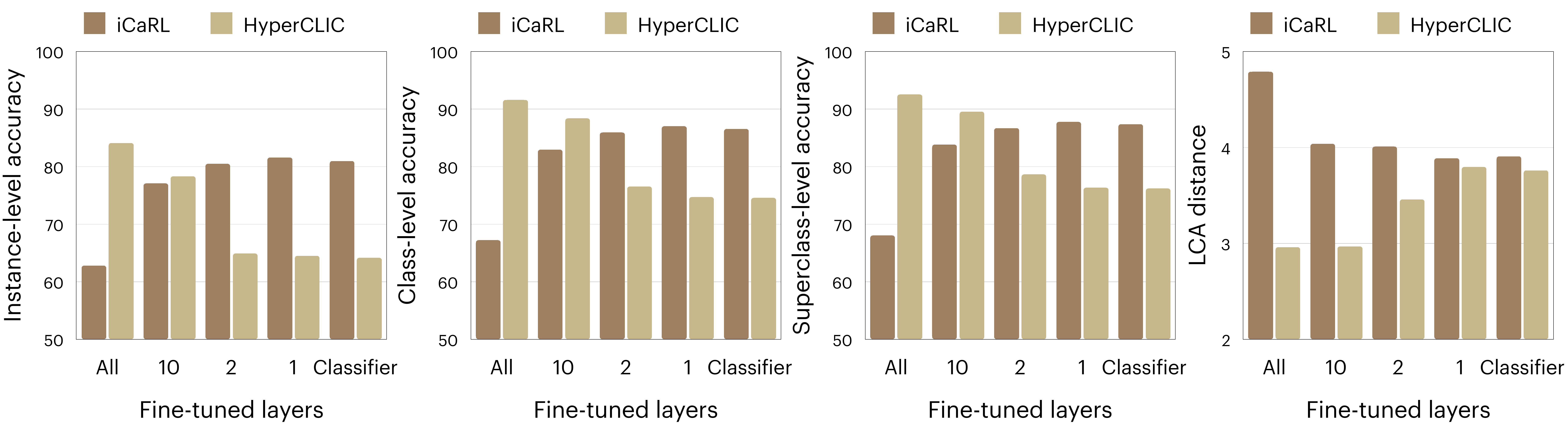}
  \caption{\textbf{Performance of HyperCLIC and iCaRL across four metrics} with varying numbers of fine-tuned layers. HyperCLIC excels when all layers are fine-tuned, while iCaRL performs better with more fixed layers. The Table \ref{tab:SOTA:egoobject:all} comparison in the pretrained scenario considers the optimal setup for each model.}
  \label{tab:finetune_n_layers:egoobject}
\end{figure}

We observed that the results of our method and iCaRL differ based on the number of layers fine-tuned in the pretrained scenario. Figure \ref{tab:finetune_n_layers:egoobject} illustrates the impact on four hierarchical metrics as the number of fine-tuned layers varies for both methods. For iCaRL, increasing the number of fine-tuned layers results in worse performance, with lower instance-, class-, and superclass-level accuracies, and higher LCA distance. Conversely, for HyperCLIC, more fine-tuned layers lead to improved results. We hypothesize that this is because \mymethod enforces a hierarchical structure on the representations, and since the pretrained model was trained in Euclidean space, fine-tuning more layers allows \mymethod to leverage the model's capacity to learn hierarchically-aware representations from the first layer. Given these trends, we use the best scores for each method to compare \mymethod with the baseline in the pretrained scenario.

\subsection{Statistical Significance} 
\label{sec:statisticall_significance}To evaluate the generality and stability of our method across different runs, we compare \mymethod with the best-performing baselines, iCaRL and DER, in both from-scratch and pretrained settings. Each experiment is conducted using 5 different random seeds, and we report the average and standard deviation of the results in Table \ref{tab:extra_runs}. Additionally, we perform statistical significance testing to compare \mymethod with the baselines. Table \ref{tab:significance_testing} presents the p-values for each metric. The results demonstrate that there are statistically significant differences between HyperCLIC and the two baselines across all metrics. Specifically, all p-values are below the commonly used threshold of $p < 0.05$, confirming that the observed performance improvements are statistically significant.

\begin{table*}[t]
\centering
\caption{\reb{\textbf{Statistical Significance Testing (p-values) of \mymethod Compared with iCaRL and DER in From-Scratch and Pretrained Settings.} \mymethod demonstrates statistical significance compared to iCaRL and DER across all metrics, with a threshold of $p < 0.05$.}} 
\label{tab:significance_testing}
\setlength{\tabcolsep}{10pt}
\reb{\begin{tabular}{@{}l *{5}{c}@{}}
\toprule
& \multicolumn{5}{c}{\textbf{From scratch}} \\
\cmidrule(lr){2-6}
& {Instance} & {Class} & {Superclass} & {LCA} & {Forgetting} \\
\midrule
iCaRL  & 3.80e-7 & 1.46e-7 & 1.45e-7 & 1.19e-9 & 1.59e-4 \\ 
DER    & 1.53e-6 & 5.79e-7 & 3.96e-7 & 2.71e-6 & 2.84e-10  \\
\midrule
& \multicolumn{5}{c}{\textbf{Pretrained}} \\
\cmidrule(lr){2-6}
& {Instance} & {Class} & {Superclass} & {LCA} & {Forgetting} \\\midrule
iCaRL& 2.84e-2 & 1.61e-5 & 4.98e-6 & 1.76e-11 & 2.60e-2 \\
DER & 4.50e-2 & 9.29e-6 & 4.21e-6 & 2.07e-7 & 3.00e-3\\
\bottomrule
\end{tabular}}
\end{table*}

\paragraph{Extra Runs} In continual learning, the order of class presentation can significantly affect the final results. To demonstrate the generality of our method, we run \mymethod and the best-performing baselines with 5 different random seeds in both from-scratch and pretrained settings on the EgoObjects benchmark, and report the average and standard deviations. Table \ref{tab:extra_runs} presents the results of this experiment. The findings suggest that \mymethod outperforms both DER and iCaRL in all metrics across both settings.

\begin{table}[htbp]
\centering
\begingroup 
\setlength{\tabcolsep}{2.5pt}
\caption{\reb{\textbf{The average $\pm$ standard deviation of \mymethod, iCaRL, and DER, evaluated with 5 different random seeds in both from-scratch and pretrained settings on the EgoObjects benchmark.} \mymethod outperforms both strong baselines across all metrics.}}
\label{tab:extra_runs}
\reb{\begin{tabular}{lcccccc}
\toprule
 & \textbf{Pretrain} & \multicolumn{3}{c}{\textbf{Accuracy $\uparrow$}} & \textbf{LCA $\downarrow$} & \textbf{Forgetting $\downarrow$}\\
                       & & Instance & Class & Superclass & & \\ \midrule
 iCaRL & & 21.49 $\pm$ 0.91 & 22.51 $\pm$ 0.87 & 23.62 $\pm$ 0.85 & 5.45 $\pm$ 0.01 & 8.72 $\pm$ 1.15 \\
 DER & & 18.69 $\pm$ 2.63 & 19.89  $\pm$ 2.58 & 20.56 $\pm$ 2.47 & 5.46 $\pm$ 0.10 & 33.90 $\pm$ 1.18 \\
\rowcolor{Gray} \mymethod & & 38.55 $\pm$ 2.37 &  42.74 $\pm$ 2.50 & 44.68 $\pm$ 2.63 & 4.91 $\pm$ 0.03 & 3.34 $\pm$ 1.39  \\
\midrule
iCaRL &\checkmark  &81.63 $\pm$ 0.20 &	87.03 $\pm$ 0.15 &	87.77 $\pm$ 0.16&	3.91 $\pm$ 0.01 & 3.74 $\pm$ 0.30\\
DER & \checkmark &80.73 $\pm$ 0.59 &85.37 $\pm$ 0.83 &86.13 $\pm$ 0.80 &4.15 $\pm$ 0.16&9.98 $\pm$ 0.53\\
\rowcolor{Gray} \mymethod & \checkmark &83.14 $\pm$  1.24&90.83 $\pm$ 0.91 &91.75 $\pm$ 0.81& 2.94 $\pm$ 0.03&5.61 $\pm$ 1.5\\
\bottomrule
\end{tabular}}
\endgroup
\end{table}

\end{document}